\documentclass[runningheads]{llncs}

\usepackage{graphicx}
\usepackage{algorithm2e}
\usepackage{listings}
\usepackage{comment}
\usepackage{booktabs}

\usepackage{tabularx}
\usepackage{multirow}
\usepackage{url}
\usepackage[colorinlistoftodos]{todonotes}
\begin{document}
%


\title{Event Extraction for Portuguese: \\ A QA-driven Approach using ACE-2005}

%
%
\author{Luís Filipe Cunha\inst{1,2,4} \and
Ricardo Campos\inst{1,3} \and
Alípio Jorge\inst{1,2}}
%
\authorrunning{L. F. Cunha et al.}
%
\institute{LIAAD-INESC TEC, Portugal \and
FCUP-University of Porto, Portugal \and
University of Beira Interior, Portugal \and University of Minho, Portugal\\
\email{lfc@di.uminho.pt, ricardo.campos@ubi.pt, amjorge@fc.up.pt}\\
\url{http://www.inesctec.pt}
}
\maketitle              
\begin{abstract}

Event extraction is an Information Retrieval task that commonly consists of identifying the central word for the event (trigger) and the event's arguments. This task has been extensively studied for English but lags behind for Portuguese, partly due to the lack of task-specific annotated corpora.  This paper proposes a framework in which two separated BERT-based models were fine-tuned to identify and classify events in Portuguese documents. We decompose this task into two sub-tasks. Firstly, we use a token classification model to detect event triggers. To extract event arguments, we train a Question Answering model that queries the triggers about their corresponding event argument roles. Given the lack of event annotated corpora in Portuguese, we translated the original version of the ACE-2005 dataset (a reference in the field) into Portuguese, producing a new corpus for Portuguese event extraction. To accomplish this, we developed an automatic translation pipeline. Our framework obtains F1 marks of 64.4 for trigger classification and 46.7 for argument classification setting, thus a new state of the art reference for these tasks in Portuguese.


\end{abstract}
\section{Introduction}\label{sec:intro}

Over the years, event extraction has been extensively studied and found to be a difficult information extraction task \cite{li-etal-2013-joint}. It aims to extract structured data regarding ``something that happens" in a text, often understood as a specific occurrence involving one or more participants. According to the Automatic Content Extraction (ACE) 2005 annotation guidelines \cite{aceguidelines}, this involves an event mention, trigger, type, argument and corresponding role:

\begin{itemize}
    \item \textbf{Event mention}: a phrase or sentence in which an event occurs, including one trigger and an arbitrary number of arguments.
    \item \textbf{Event trigger}: the word that expresses an event occurrence.
    \item \textbf{Event type}: represents a high-level categorization of events based on their general semantic meaning. It can be composed of sub-types that provide a more specific categorization of events.
    \item \textbf{Event argument}: an entity mention, temporal expression or value that serves as a participant or attribute with a specific role in an event mention.
    \item \textbf{Argument role}: indicates the semantic relationship of the argument within the event, such as the agent that performs the action, the time or location of the event, etc.

\end{itemize}

Typically, event mentions consist of an event trigger and their corresponding event arguments. Consider the following sentence, which illustrates the process of automatically identifying and classifying the event triggers and their corresponding arguments found in the text. 

\begin{center}
\textit{``Elvis Presley morreu de ataque cardíaco em 1977, Memphis, Tennessee."}
(Elvis Presley died of a heart attack in 1977, Memphis, Tennessee.)
\end{center}
In this example, the word \textit{``morreu"} (died) is an event trigger of type \texttt{Life.Die} and ``Elvis Presley" is an event argument with role \texttt{Victim}.

While several English event extraction systems already exist \cite{BALALI2020106492,nguyen-etal-2016-joint-event,yang-mitchell-2016-joint}, they reveal poor portability to other languages due to dependencies on English annotated textual resources. In this paper, we aim to tackle this problem in the context of the Portuguese language.
In particular, we aim to develop a method that allows for the extraction of event mentions by leveraging the power of Transformers-based models \cite{vaswani2017attention}. To address this problem, we divided the event extraction task into two sub-tasks: Trigger extraction and Argument Extraction.

We approach trigger identification and classification as a Token Classification task. Then for the Argument extraction, we use a Question Answering (QA) model (inspired by Du et al. \cite{Du2020}) where we question the event trigger about its corresponding event argument roles. To perform these tasks, we used BERTimbau \cite{souza2020bertimbau}, a  BERT \cite{DBLP:journals/corr/abs-1810-04805} model pre-trained on Portuguese textual data. We fine-tuned this model with event annotations from a Portuguese translated version of ACE-2005 \cite{doddington-etal-2004-automatic}, containing textual data annotated with event triggers and corresponding arguments. For the QA task, we also experimented with the SQuAD \cite{Rajpurkar2016} dataset in order to train our model to perform extractive QA.

Since ACE-2005 was not available in Portuguese, and the Portuguese annotated corpora we found \cite{duran2012propbank,Costa:Branco:2010} did not contain explicit annotations for both event triggers and arguments, we decided to automatically translate the ACE-2005 corpus from English to Portuguese. For this purpose, we developed a translation pipeline that enabled us to automate the translation and alignment tasks. This translated dataset is an important contribution to this current work.

The main \textbf{contributions} of this work are listed below:

\begin{itemize}
    \item A pipeline for dataset translation and annotation alignment that allows the translation of annotated datasets to the Portuguese language.

    \item Based on this pipeline, we produced a new dataset by translating ACE 2005 for Portuguese, which is already in the process of being accepted at the Linguistic Data Consortium (LDC) repository. 

    \item Using the Portuguese version of ACE-2005 corpus, we produced and deployed event extraction models. These models correspond to a baseline for Portuguese event extraction.

    \item Based on the produced models, we developed and made available on Huggingface Hub, an event extraction framework for the Portuguese language. 
\end{itemize}

\section{Related Work}

Event extraction is a fundamental task in Natural Language Processing (NLP) that has been widely researched in recent years mainly for English and with less attention to other languages. Over the years, several approaches have been proposed to tackle this task, ranging from traditional rule-based methods to more advanced machine learning and deep learning techniques \cite{lai2022event,li2022survey}.

Recent works Du et al. \cite{Du2020} have demonstrated promising results using QA models in event extraction. The authors leveraged BERT \cite{DBLP:journals/corr/abs-1810-04805} models fine-tuned on the ACE-2005 corpus, to identify event triggers and corresponding arguments.

Huang et al. \cite{huang-etal-2018-zero} addressed event extraction by using Zero-Shot Learning to handle the scarcity of annotated data and the limited range of event types, which constrains the applicability of this task to certain domains. They drew inspiration from Pustejovsky et al. \cite{PUSTEJOVSKY199147}, who proposed that the semantics of an event structure can be generalized to different event mentions. Following this idea, they used an event ontology that defines structures for each event type. The authors used Abstract Meaning Representations (AMR) \cite{banarescu-etal-2013-abstract} to identify the event triggers and argument candidates, constructing a structure for each event.


Although event argument extraction has been primarily approached as a sentence-level task, it should be noted that in real-world scenarios, the arguments of an event can be dispersed across multiple sentences. To address this problem, Li et al. \cite{Li-Ji-Han-2021-event} propose a document-level approach for argument extraction. They use a generative model (BART \cite{lewis2019bart}, T5 \cite{raffel2020exploring}) that is conditioned by the input sequence and unfilled templates created from an event ontology. The model is responsible for filling those templates with a limited vocabulary in order to prevent it from ``hallucinating".

Despite the advances in this particular area, little has been done for the Portuguese language. Quaresma et al. \cite{info10060205} implemented an Event Extraction framework for the Portuguese language, focused on the crime investigation domain. Their framework relied on Semantic Role Labeling to extract event arguments and was validated on PropBank \cite{duran2012propbank} corpus. The authors did not classify events by type and instead focused on the roles provided by the SRL schema, such as Actor, Place, Time, and Object. Consequently, it would be difficult to compare their work with ours as ACE-2005 contains 33 different event types. 

The same applies to the work developed by Costa and Branco \cite{Costa2012}. They used feature engineering combined with a decision tree trained on the TimeBankPT corpus to extract events from Portuguese texts. However, the TimeBankPT corpus event annotations only contain the following event types: REPORTING, OCCURRENCE, STATE, I\_STATE, and I\_ACTION. These annotations lack detail on event structure compared to ACE-2005, specifically the event arguments and roles.


\section{Methodology}

This section will discuss the methodologies used to extract event triggers (Section \ref{sec:meth_trigger}) and event arguments (Section \ref{sec:meth_argument}). To achieve this, we fine-tune a Portuguese BERT model \cite{souza2020bertimbau} with a Portuguese-translated version of the ACE-2005 corpus (more details in Section \ref{sec:translate}). We fine-tined the model for token classification and Question Answer tasks to extract event triggers and event arguments, respectively.

\subsection{Trigger Extraction}\label{sec:meth_trigger}

For the first task, we train a model that identifies and classifies event triggers. The task is formulated as a token classification one. Given a sequence of $N_1$ tokens \(W = [w1, w2, ..., wN_1 ]\) and a fixed set of event types (None type included) of length $N_2$ \(A = [a1, a2, ..., aN_2 ]\)  our model assigns each token from $W$ to their corresponding label from $A$.

To perform this task we used BERTimbau \cite{souza2020bertimbau}, a BERT-based model that was pre-trained on Portuguese texts. We fine-tuned this model on token classification using the Portuguese-translated version of the ACE-2005 corpus. For that, we converted the translated corpus to the IOB scheme \cite{ramshaw-marcus-1995-text} (\textbf{B}eginning, \textbf{I}nside and \textbf{O}utside) where a label is assigned to each token of the text sequence. We consider the 9 event types and 33 event sub-types contained in ACE-2005 as labels for token classification task.

\subsection{Argument Extraction}\label{sec:meth_argument}

To extract arguments from the text, we used extractive QA, where we formulated questions about the event to obtain the argument roles. These questions are influenced by each specific trigger word. Given a sentence $S$ and a question $Q$, this task aims to find the token span offsets where the corresponding answer $a$ lies in $S$. In order to accomplish this objective, we fine-tuned the BERTimbau model in a QA task. 

The input sequence format is described below:

\[ \mathbf{[CLS]\ question (Q)\ [SEP]\ sentence\ (S) [SEP]}\]

In this format, we have the BERT classification token \texttt{CLS} and the \texttt{SEP} token that separates the $S$ and $Q$ input text sequences. The model outputs logits for the start ($a_{start}$) and end ($a_{end}$) positions of the answer to each token of the input sequence. Before selecting the most probable answer offsets, several validations must be performed to ensure that the answer span is valid. For instance, the answer $a$ should be fully contained within the sentence $S$ and not part of the question $Q$; The start offset $a_{start}$ cannot be greater than the end offset $a_{end}$, etc. These validations are common procedures in the QA task.

\subsubsection{Questions Generation}

In the following, we outline how we generated the actual questions for fine-tuning the model. We adopted a template-based approach, similar to Du et al. \cite{Du2020} and Lyu et al. \cite{lyu-etal-2021-zero}. Based on the event type, we can determine the appropriate questions to ask in order to extract specific arguments. In ACE-2005, each event type has a predetermined set of argument roles. We generated a question template for each event type by creating a set of questions (in Portuguese) based on the event type's corresponding roles. Each question of the template expects to obtain as an answer the argument associated with each role. 
We referred to the argument roles description provided in the ACE-2005 annotation guidelines to generate these questions.

Table \ref{tab:QTemplates} contains the questions used to extract the arguments of an event type \texttt{LIFE.DIE}. Following the guidelines \cite{aceguidelines}, we know that this event type can have five different argument roles: Agent, Victim, Instrument, Time and Place.

\begin{table}[]
\centering
\caption{Question templates for the event type LIFE.DIE.}\label{tab:QTemplates}
\begin{tabular}{@{}lll@{}}
\toprule
\textbf{Role}                   & \textbf{Question (Portuguese)}                                & \textbf{Question (English)}  \\ \midrule
\multicolumn{1}{l|}{Agent}      & \multicolumn{1}{l|}{\textit{Quem é o assassino?}}             & Who is the assassin?         \\
\multicolumn{1}{l|}{Victim}     & \multicolumn{1}{l|}{\textit{Quem morre?}}                     & Who died?                    \\
\multicolumn{1}{l|}{Instrument} & \multicolumn{1}{l|}{\textit{Qual é o instrumento utilizado?}} & What is the used instrument? \\
\multicolumn{1}{l|}{Time}       & \multicolumn{1}{l|}{\textit{Quando ocorre a morte?}}          & When is the death?           \\
\multicolumn{1}{l|}{Place}      & \multicolumn{1}{l|}{\textit{Onde ocorre a morte?}}            & Where is the death?          \\ \bottomrule
\end{tabular}
\vspace{-\baselineskip}
\end{table}

Then, to contextualize the question within the event mention, we concatenate it with the event trigger word, a method that has shown to improve the model results \cite{Du2020}. We use the following question format: \texttt{\{question\} + in \{trigger\}?}. For instance, in our example of Section \ref{sec:intro}, we have an  event of type \texttt{LIFE.DIE}. In order to extract the argument role \texttt{Time}, the following question is generated:
\begin{center}
    \textit{Quando ocorre a morte + em morreu?}\\
    (When is the death + in died?)
\end{center}
Given this prompt, the model should output the answer span corresponding to ``\textit{em 1977}" (in 1997). The model uses the generated questions to extract each argument role from the text. Given an event mention and an event trigger, we replicate this procedure for all the event arguments.

\subsubsection{Impossible Answer}

It's important for the model to be able to identify questions that do not have a correct answer. In fact, not all event argument roles can be found in every event mention. For instance, in the example provided, the \texttt{Agent} argument role cannot be found in the text, which implies that the question ``Who is the assassin?" should not have a correct answer.

To address this problem, we trained the model to predict the ``impossible" answer. During the training phase, we gave the model several questions without any answer. In these cases, the answer span offsets are assigned to the index 0 of the input sequence corresponding to the BERT CLS token. By doing so, during inference, our model is able to filter out the roles that may not be present in the text by giving the highest probability to the CLS token. In that case, we consider that the argument role is not present in the current event mention.

\section{Data}\label{sec:data}

When it comes to the event extraction task, ACE-2005 is considered the standard corpus in this field. While other corpora such as PropBank focus on the annotation of predicate-argument structure, ACE-2005 goes beyond this by providing information on the overall event structure, including the event type and its corresponding argument roles. It is available in English, Chinese, and Arabic, however, there is no version of this dataset in Portuguese. We decided to take the effort of translating the dataset, thus being able to work with this valuable resource for Portuguese. In this work, we used a translated version of ACE-2005 in Portuguese, which contains 5 526 event mentions consisting of 5 526 event triggers and 9 649 corresponding event arguments.

 We have also used the well known SQuAD corpus \cite{Rajpurkar2016} for training an extractive question answering model. It consists of articles obtained from Wikipedia and a set of corresponding questions and answers about each article. In this work, we used two versions of this dataset: SQuAD1.1, which contains 100 000 questions and respective answers; SQuAD2.0 \cite{squad2}, which contains 150 000 questions and answers. The latter version contains all the questions from version 1.0, however, it contains 50 000 additional questions that have no correct answers. In version 2.0, one must consider the impossible answer scenario when finding the correct answer, creating a more challenging QA task. A Portuguese version of SQuAD 1.0 was already available from the Deep Learning Brasil Group\footnote{http://www.deeplearningbrasil.com.br/
}, however, we took the effort of translating version 2.0.

\subsection{ACE-2005 Translation}\label{sec:translate}

In this section, we provide an overview of the ACE-2005 corpus translation process. Although we use automatic translation, translating an annotated dataset can become particularly challenging when it comes to aligning its annotations. In fact, mismatches can occur between the annotations and their occurrences in the corresponding sentence. For instance, in sentence ``The troops land on the shore", ACE-2005 states that the trigger ``land" should be extracted. However, the word ``land" is translated to ``\textit{terra}" (land as a noun) in isolation and to ``\textit{desembarcam}" (land as a verb) in context.

In the pre-processing\footnote{https://github.com/nlpcl-lab/ace2005-preprocessing} of ACE-2005, each event annotation was assigned to its corresponding text sentence. Then, we automatically translated each sentence, its corresponding triggers and arguments. These translations resulted in annotations' miss-alignments i.e., translated annotations that were not contained in the translated sentences. In order to align these cases, we developed an alignment pipeline that is composed of four components: lemmatization, multiple translations, a BERT-based world aligner and fuzzy string similarity. 

Regarding lemmatization, instead of directly matching the annotations to their respective sentence, we calculated lemma tokens from both the translated sentence and the corresponding translated annotations. Then, we performed the matching process using these lemma tokens. If that method was not able to find a match, we proceed to the next element of the pipeline. In particular, we used Microsoft Dictionary Lookup API to retrieve alternative translations of the event annotations and tried to match them in their sentences.

The third component of our pipeline involved aligning the annotations with a parallel corpus word aligner, proposed in Dou et al. \cite{Dou2021}. In practice, we used the embeddings retrieved from the BERT-Multilingual model \cite{DBLP:journals/corr/abs-1810-04805} in order to compute the correspondence between each token of the source sentences (English) and the translated sentences (Portuguese). Then by looking at the English annotations words, we calculated the corresponding Portuguese annotations.

Finally, if the previous approaches could not solve the miss-alignment, we used character-level similarity algorithms such as Levenshtein distance \cite{levenshtein1965} and  Gestalt pattern matching \cite{ratcliff1988pattern}. This approach allowed us to identify the substring within the sentence that was most similar to the annotations.


\subsection{SQuAD Translation}
In this work, we used a version of SQuAD1.1 that had been previously translated into Portuguese. To create a Portuguese version of the SQuAD2.0 dataset, we automatically translated the additional 50,000 impossible questions. Since these questions do not have a valid answer within the article's text, performing annotation alignments to this dataset was unnecessary.

\section{Modeling}

To validate our approach we use the translated ACE-2005 corpus for training and testing, as well as the translated SQuAD datasets for modeling question-answering. We aim to assess the following: 1) the success of the trigger identification and classification approach; 2) the success of the question answering approach for argument classification; 3) the impact of training the model to detect the absence of event arguments. Given the lack of other works in Portuguese,  we compare our work with the results obtained by state of the art approaches for the same tasks on the original ACE-2005 corpus.

Our first setup was to fine-tune the BERTimbau model \cite{souza2020bertimbau} with the train split from our translated version of ACE-2005 (BERT-PT-ACE05). We used the event trigger annotations to train the token classification model and the argument annotations to train the QA model.

Then, for the argument extraction task, we used an existing Portuguese QA model \cite{pierreguillou2021bertbasecasedsquadv11portuguese} (pre-trained on SQuAD1.1 dataset ) and fine-tune it with the ACE-2005 data (BERT-PT-SQuAD1.1-ACE05). The motivation of this approach consisted of teaching the model to answer general questions first and then using that knowledge to answer our event-driven questions to extract the event arguments. Due to the nature of the SQuAD1.1 dataset, the Portuguese QA model \cite{pierreguillou2021bertbasecasedsquadv11portuguese} is not able to output impossible answers. 

Next, we tested a similar approach, but instead of using a QA model based on SQuAD1.1, we fine-tuned the BERTimbau model with SQuAD2.0 so the model could learn how to identify impossible answers. Subsequently, we used ACE-2005 data so the model learns how to extract the event arguments (BERT-PT-SQuAD2.0-ACE05). As stated before, dealing with impossible answers is important because not all event argument roles are present in every event.

\section{Results}
Our models were validated on the test split of ACE-2005 containing 422 event triggers and 892 arguments. We ensured a fairer comparison with state-of-the-art in English by using the same data splits and evaluation criteria as previous works \cite{Du2020,li-etal-2013-joint}. A correct identification and classification of an event trigger requires matching its offsets and event type with the gold-standard. An event argument's correct identification and classification demands matching its offsets with the ACE-2005 annotations and ensuring its semantic role is accurate.
In other words, matching between extracted elements and ground truth must be exact.

\begin{table}[]

\caption{Event Extraction results on ACE-2005 dataset.}
\label{tab:results}
\begin{tabularx}{\textwidth}{XXXXXXX}
\toprule
\multicolumn{1}{c}{\multirow{2}{*}{\textbf{Model}}}                    & \multicolumn{3}{c}{\textbf{Trigger Classification}}           & \multicolumn{3}{c}{\textbf{Argument Classification}} \\ \cmidrule(l){2-7} 
\multicolumn{1}{c}{}                                                   & \multicolumn{1}{|c}{\textbf{P}} & \multicolumn{1}{c}{\textbf{R}} & \multicolumn{1}{c}{\textbf{F1}} & \multicolumn{1}{|c}{\textbf{P}}     & \multicolumn{1}{c}{\textbf{R}}     & \multicolumn{1}{c}{\textbf{F1}}     \\ \midrule

\multicolumn{7}{c}{\textbf{English ACE-2005}} \\ \midrule
\multicolumn{1}{l|}{JRNN  2016 \cite{nguyen-etal-2016-joint-event}}        & \multicolumn{1}{c}{73.0}       & \multicolumn{1}{c}{66.0}       & \multicolumn{1}{c|}{69.3}       & \multicolumn{1}{c}{56.7}           & \multicolumn{1}{c}{54.2}           & \multicolumn{1}{c}{55.4}            \\
\multicolumn{1}{l|}{JointEntityEvent 2016 \cite{yang-mitchell-2016-joint}} & \multicolumn{1}{c}{75.1}       & \multicolumn{1}{c}{63.3}       & \multicolumn{1}{c|}{68.7}       & \multicolumn{1}{c}{70.6}           & \multicolumn{1}{c}{36.9}           & \multicolumn{1}{c}{48.4}           \\
\multicolumn{1}{l|}{GAIL-ELMo  2019 \cite{10.1162/dint_a_00014}}            & \multicolumn{1}{c}{74.8}       & \multicolumn{1}{c}{69.4}       & \multicolumn{1}{c|}{72.0}       & \multicolumn{1}{c}{61.6}           & \multicolumn{1}{c}{45.7}           & \multicolumn{1}{c}{52.4}            \\ 
\multicolumn{1}{l|}{BERT\_QA\_Arg   2020      \cite{Du2020}}           & \multicolumn{1}{c}{71.1}       & \multicolumn{1}{c}{73.7}       & \multicolumn{1}{c|}{72.4}       & \multicolumn{1}{c}{56.8}           & \multicolumn{1}{c}{50.2}           & \multicolumn{1}{c}{53.3}            \\
\multicolumn{1}{l|}{OneIE 2020  \cite{lin-etal-2020-joint}} & \multicolumn{1}{c}{-}          & \multicolumn{1}{c}{-}          & \multicolumn{1}{c|}{74.7}       &            \multicolumn{1}{c}{-}    &     \multicolumn{1}{c}{-}           & \multicolumn{1}{c}{56.8}            \\
\multicolumn{1}{l|}{Text2Event   2021 \cite{lu-etal-2021-text2event}}  & \multicolumn{1}{c}{69.6}          & \multicolumn{1}{c}{74.4}          & \multicolumn{1}{c|}{71.9}       &      \multicolumn{1}{c}{52.5}          &         \multicolumn{1}{c}{55.2}       & \multicolumn{1}{c}{53.8}            \\
\multicolumn{1}{l|}{FourIE   2021 \cite{nguyen-etal-2021-cross}}       & \multicolumn{1}{c}{-}          & \multicolumn{1}{c}{-}          & \multicolumn{1}{c|}{75.4}      &      \multicolumn{1}{c}{-}          &      \multicolumn{1}{c}{-}          & \multicolumn{1}{c}{58.0}              \\
\multicolumn{1}{l|}{GraphIE 2022 \cite{nguyen-etal-2022-joint}}        & \multicolumn{1}{c}{-}          & \multicolumn{1}{c}{-}          & \multicolumn{1}{c|}{\textbf{75.7}}      &       \multicolumn{1}{c}{-}         &      \multicolumn{1}{c}{-}          & \multicolumn{1}{c}{\textbf{59.4}}            \\ \midrule

\multicolumn{7}{c}{\textbf{Portuguese ACE-2005}} \\ \midrule
\multicolumn{1}{l|}{BERT-PT-ACE05}                                              & \multicolumn{1}{c}{63.6}       & \multicolumn{1}{c}{65.3}       &  \multicolumn{1}{c|}{64.4}       &           \multicolumn{1}{c}{46.3}      &         \multicolumn{1}{c}{45.1}        &       \multicolumn{1}{c}{45.7}           \\
\multicolumn{1}{l|}{BERT-PT-SQuAD1.1-ACE05}                                                  &     \multicolumn{1}{c}{-}       &       \multicolumn{1}{c}{-}     &     \multicolumn{1}{c|}{-}      &     \multicolumn{1}{c}{45.7}           &      \multicolumn{1}{c}{46.3}          &    \multicolumn{1}{c}{46.0}             \\ 
\multicolumn{1}{l|}{BERT-PT-SQuAD2.0-ACE05}                                                  &     \multicolumn{1}{c}{-}       &       \multicolumn{1}{c}{-}     &     \multicolumn{1}{c|}{-}      &          \multicolumn{1}{c}{46.8}         &       \multicolumn{1}{c}{46.6}          &     \multicolumn{1}{c}{\textbf{46.7}}           \\ \bottomrule
\multicolumn{1}{l}{}                                                  &            &         &          &                         \\
\end{tabularx}
 \vspace{-\baselineskip}
\end{table}


On Table \ref{tab:results} is presented a comparison of our results using the ACE in Portuguese against SOTA event extraction methods using the original ACE 2005. The evaluation metrics are Precision (P), Recall (R) and F1 scores. 
Looking at the trigger extraction task, our F1 for Portuguese is about 10\% below F1 for English. This superiority, also observable for argument extraction (nearly 15\%), is probably due to the translation effects and language specifics. As for the arguments, one can observe a slightly positive impact of using question-answering and a clearer impact of allowing non-answers (more details in Section \ref{sec:discussion}). 

Finally, for direct qualitative evaluation, we developed and deployed a Web application\footnote{https://hf.co/spaces/lfcc/Event-Extractor} that functions as an interface  for the proposed event extraction framework making our models accessible and usable.

\section{Discussion}\label{sec:discussion}

Our models for Portuguese were trained using automatically translated data. However, the automatic translation still faces many challenges, including accurately capturing the nuances of language, handling idiomatic expressions, and dealing with cultural language differences. Therefore, it is important to be aware of these limitations and expect some noise in the translated data.

Another limitation we found was the annotation alignment. In fact, we used several techniques to improve our results in the alignment of the trigger and argument span annotations. Despite that, we know there are still alignment errors, causing triggers and arguments to be wrongly annotated. Consider the following sentence ``We discussed the Middle East peace process" and the corresponding translation ``\textit{Discutimos o processo de paz no Médio Oriente}". In this sentence, the word ``discussed" is an event trigger of type \texttt{Contact.Meet} while the word ``We" corresponds to an event argument playing the role \texttt{Entity}. However, in the Portuguese translation, the sequence ``We discussed" was translated into ``\textit{Discutimos}" (the verb was conjugated in the first person plural). The argument ``We" became implicit, making the annotation hard to align.

Furthermore, in addition to the translation noise, we believe that the event extraction difficulty for English and Portuguese languages is not the same. For instance, the Portuguese language has a greater diversity of words. This is the case of the conjugation of verbs. Looking at the trigger words, the ACE-2005 corpus has about 1237 different trigger words in total, while the Portuguese translated version has 1900 trigger words. Although we show comparative results of our work against SOTA English models, it is not entirely fair to make a direct comparison given the differences in language and cultural context.

As for the results, our validation data was translated in the exact same manner as our training data, which means that it also contains the translation and alignment noise we mentioned above. It would be interesting to validate our models against data revised by humans, ensuring a higher data quality.

In fact, we employed identical metrics as previous works to compare our outcomes. Nonetheless, the strict evaluation metrics hide many near misses. Consider the following example: 
\begin{center}
\ \ \ \ \textbf{Gold Argument:} \textit{ex-banqueiro sênior Callum McCarthy} \\
    \textbf{Predicted Argument:} \textit{O ex-banqueiro sênior Callum McCarthy} \\
    (The former senior banker Callum McCarthy) \\
\end{center}
In this case, our model prediction closely matches the ground truth but receives no credit for it because it fails to identify the determinant ``O" (the).

Finally, another limitation of our method is that ACE-2005 annotations are sentence-level, causing our models to have difficulties extracting cross-sentence event arguments. In order to attenuate this problem, our deployed framework uses a context window that works as a hyperparameter allowing us to consider more than one sentence as context for the QA task.

\section{Conclusion}


This work proposes a novel method for extracting events from Portuguese text. Our approach involves two tasks: first, we classify and identify event triggers using token classification; Then, we extract event arguments using extractive QA. To train models capable of performing those tasks, we fine-tune the BERTimbau model with SQuAD and ACE-2005 datasets, the latter being a reference in the event extraction field. Since these datasets were not available in Portuguese, we developed a translation pipeline to automatically translate them. We present a new event extraction baseline for Portuguese using the ACE-2005 dataset translated into Portuguese. As we could not find any comparable works in Portuguese, we used existing English event extraction works as a benchmark.  While our models achieved lower results compared to the English models, we believe the comparison cannot be made directly due to language differences.

For future work, considering the lack of extensive research on this task for Portuguese, there are numerous suitable approaches that could improve our results. For example, expanding our data domain by incorporating other event-driven datasets, such as TAC KBP 2015 \cite{kbp2015} and MINION \cite{minion}. We could also leverage Semantic Role Labeling for Portuguese in order to enhance the performance of event argument extraction.

\section{Acknowledgments}
The authors of this paper were financed by National Funds through the FCT - Fundação para a Ciência e a Tecnologia, I.P. (Portuguese Foundation for Science and Technology) within the project StorySense, with reference 2022.09312.PTDC)

%
%
%
 \bibliographystyle{splncs04}
 \bibliography{refs}

\end{document}